\documentclass{article}

\usepackage[final]{machine_learning_and_the_physical_sciences_workshop}

\usepackage[T1]{fontenc}    % use 8-bit T1 fonts
\usepackage{hyperref}       % hyperlinks
\usepackage{url}            % simple URL typesetting
\usepackage{booktabs}       % professional-quality tables
\usepackage{amsfonts}       % blackboard math symbols
\usepackage{nicefrac}       % compact symbols for 1/2, etc.
\usepackage{microtype}      % microtypography

\usepackage{xcolor}         % colors
\usepackage{enumerate}

\usepackage{amsmath}
\usepackage{multirow}
\usepackage{graphicx}
\usepackage{bbm}
\usepackage{amssymb}
\usepackage{graphics}
\usepackage{makecell}

\title{Time Series Viewmakers for Robust Disruption Prediction}

\author{%
  Dhruva Chayapathy \\
  Alpharetta High School \\
  \texttt{dhruva.chayapathy@gmail.com} \\
  \And
  Tavis Siebert \\
  UC Berkeley \\
  \texttt{tsiebert@berkeley.edu} \\
  \AND
  Lucas Spangher \\
  Google Research \\
  \texttt{spangher@google.com} \\
  \And
  Akshata Kishore Moharir \\
  Microsoft \\
  \texttt{akshatan@terpmail.umd.edu} \\
  \And
  Om Manoj Patil \\
  Alpharetta High School \\
  \texttt{om1.patil.om@gmail.com} \\
  \And
  Cristina Rea \\
  MIT Plasma Science and Fusion Center\\
  \texttt{crea@psfc.mit.edu} \\
}
\begin{document}

\maketitle

\begin{abstract}
  Machine Learning guided data augmentation may support the development of technologies in the physical sciences, such as nuclear fusion tokamaks. Here we endeavor to study the problem of detecting disruptions — i.e. plasma instabilities that can cause significant damages, impairing the reliability and efficiency required for their real world viability. Machine learning (ML) prediction models have shown promise in detecting disruptions for specific tokamaks, but they often struggle in generalizing to the diverse characteristics and dynamics of different machines. This limits the effectiveness of ML models across different tokamak designs and operating conditions, which is a critical barrier to scaling fusion technology. Given the success of data augmentation in improving model robustness and generalizability in other fields, this study explores the use of a novel time series viewmaker network to generate diverse augmentations or "views" of training data. Our results show that incorporating views during training improves AUC and F2 scores on DisruptionBench tasks compared to standard or no augmentations. This approach represents a promising step towards developing more broadly applicable ML models for disruption avoidance, which is essential for advancing fusion technology and, ultimately, addressing climate change through reliable and sustainable energy production.
\end{abstract}

\section{Introduction}

One of the most current notable efforts to get an idea from physical sciences into commercializable reality is nuclear fusion. Nuclear fusion has long been considered a ``holy grail'' in providing carbon-free energy without significant waste or land demands \citep{GIULIANI2023113554, spangher2019characterizing}. A leading approach to generating fusion in laboratory plasmas is \textit{magnetic confinement} via tokamak machines. However, the commercial viability of tokamaks hinges on developing their capability to accurately predict plasma disruptions: losses in plasma stability which may damage a tokamak and take it offline for months at a time.

Disruption are difficult to predict with first principles or physics-only models. Thus, a significant focus within the nuclear fusion community has been on the development of machine learning (ML) models for disruption prediction. ML models are already in production at major tokamaks, including random forest predictors and simple neural networks \citep{diiidRea, eastHu, asdexDis}, and research into various deep learning architectures has established the state of the art in disruption prediction capabilities \citep{zhuHybridDeeplearningArchitecture2020, arnold2023continuous, spangher2023autoregressive, zhuHybridDeeplearningArchitecture2020}. However, because the characteristics and physical dynamics of a tokamak can change dramatically depending on the machine's design
\citep{granetz2018machine}, and over time as the machine is upgraded \citep{asdexUpgrade} or modified \citep{cMod}, it is imperative that prediction models can generalize to these different sources of data.

Data augmentation is one way to improve model generalizability \citep{Iwana2021} and robustness \citep{YangRobust}. Since tokamak data is comprised of multivariate time series, common augmentation strategies can include jittering, slicing, and warping \citep{Um_2017, wen2019time}. However, 
naively applying these techniques can lead to augmentations that fail to mimic plausible plasma behavior, impairing the model's ability to learn patterns relevant for disruption prediction \citep{fu2020data}. Previously, a Stutent t process based approach has been used for tokamak data \citep{dpredaug22}, however it's direct impact on post-hoc models was not evaluated. In this work, we explore the use of ``Viewmaker Networks'' \citep{tamkin2021viewmaker}, models which learn to generate augmentations through adversarial learning. We hypothesize that viewmaker augmentations, or ``views", are a solution for incorporating realistic yet diverse augmentations of disruption data which can improve the robustness of models toward disruptive discharges, and we demonstrate that using views in training which improve model AUC and F2 on DisruptionBench \citep{lucas2024disruptionbench} tasks over multiple disruption prediction models when compared to training on standard augmentations such as those present in the tsaug library or training without augmentation.

The novelty of our paper is as follows:
\\
\hspace*{4mm}
\begin{minipage}{0.9\textwidth}
    (1) We present an original adaptation of the image-based viewmaker network for time series augmentation and benchmark against other time series augmentation techniques.\\
    (2) We are the first to rigorously test generative augmentation methods in the context of nuclear fusion datasets. 
\end{minipage}

\section{Background}
\label{sec:background}

\textbf{Disruption benchmarks:} 
DisruptionBench \citep{lucas2024disruptionbench} is the first disruption prediction benchmark aimed at assessing model generalizability across tokamaks. The data is comprised of roughly 30k trials (referred to here as ``discharges'') from three tokamaks: General Atomics' DIII-D, MIT's Alcator C-MOD, and the Chinese Academy of Science's EAST. 
Training and tests sets for each task can be generated by specifying the following parameters: (1) \textbf{New Machine}: One of three tokamaks is designated as the "new machine." The test set is comprised of only "new machine" discharges while the discharges in the training set are specified by the "data-split" parameter. (2) \textbf{Data-Split}: In this work, we focus on the following four cases with relation to two training machines and one test machine: Case 1, Zero-Shot; Case 2, Few-Shot; Case 3: Many-Shot. Case 4: Train and test from a single machine. For details on how true and false positives are calculated in DisruptionBench, please refer to Appendix \ref{app:disruptionbench}

\textbf{Dataset provenance and preparation:} 
Our dataset is open\footnote{Please find our data at the following url: \url{https://dataverse.harvard.edu/dataset.xhtml?persistentId=doi:10.7910/DVN/XIOHW1}.} \citep{zhuHybridDeeplearningArchitecture2020}. Each tokamak discharge lasts from seconds to minutes, generating gigabytes of multi-modal data, of which we focus on discharges' global state variables listed in Table \ref{tab:features}.  The tokamaks may be generally characterized by the information in Table \ref{tab:datasets}, which demonstrates significant difference in average discharge length, sampling rate, and number of samples. 

Each discharge is of length $n$, and we discard discharges that are shorter than 125ms. Each tokamak's diagnostics store metrics from each discharge at a different sampling rate. We normalize across tokamaks by discretizing at 5ms and interpolating via forward-fill. Roughly 10\% of discharges end in disruptions. We truncate our time series to $X=1,...,n-\nu$, where $n$ is the full length and $\nu=40ms$, the minimum amount of time for a disruption mitigation system to activate. For a description of the features chosen, please see Table \ref{tab:features}. 

\textbf{Viewmaker networks:} 
Viewmaker Networks \citep{tamkin2021viewmaker} are one approach to domain-agnostic contrastive learning. They adversarially train an encoder network and a generative network (viewmaker) similar to a Generative Adversarial Network (GAN) set-up \citep{goodfellow2014generative} while using a SimCLR loss \citep{chen2020simple}. For a more complete description of Viewmakers, please see Appendix \ref{app:viewmaker}.

\section{Models and methods}
\label{sec:models}
\textbf{Time series viewmakers:} Inspired by series decomposition blocks in \citep{wu2022autoformer}, we first decompose our time series \(X \in \mathbb{R}^{T \times d}\) into trend-cycle and seasonal components: \(X_t = \mathrm{AvgPool}(\mathrm{Pad}(X))\), \(X_s = X - X_t\).
Because we decompose the original time series, (i.e. $X=X_s+X_t$) we use a separate generator to perturb each component. The perturbations are then combined, smoothed, constrained, and added to the original time series. See Figure \ref{fig:TSViewmaker} for an illustration of time series viewmakers.

We find the selection of generator networks $V_t$ and $V_s$ to be critical to performance. It is important that these generators generate \textbf{stochastic} and \textbf{challenging} perturbations.
To better capture long-term temporal dependencies in the data, we change the viewmaker/generator from residual blocks to an LSTM-Transformer based model similar to the LSTMFormer detailed later. Noise is concatenated between each transformer layer to ensure stochastic perturbations. Furthermore, we adopt an adversarial SimCLR loss and a distortion budget $\epsilon=0.1$. We trained our viewmaker for 10000 steps with a loss weight of 2.5 and a loss temperature of 0.05339.
However, we encourage others to experiment with different loss functions, distortion budgets, generator networks, and encoders.\footnote{Our code is public and may be found at the following url: \url{https://github.com/utterwqlnut/plasma-data-augmentation.git}}

\textbf{Tested models:} For benchmarking the effectivness of these augmentations, we experiment with 3 post-hoc models: LSTMFormer, FCN, and GPT-2. Compared to transformers, the sequential processing of LSTMs makes them particularly effective in encoding temporal relations within sequences. Therefore, the LSTMFormer first uses an LSTM layer to generate time-conscious positional encodings, similar to \citep{zeyer2019compare, lim2020temporal}. The output of the LSTM layer is then propagated through a normalization layer \citep{ba2016layer}, a four-block transformer encoder \citep{vaswani2017attention}, and a classification head. The FCN is a convolutional neural network variant first introduced in \citep{wang2016timeseriesclassificationscratch}. GPT-2 introduced for disruption prediction in \citep{lucasgpt-2} is an autoregressive transformer decoder model which uses a pretraining and a curriculum training scheme.

Our model training occurred on a shared cluster of 48 Nvidia A100 GPUs through the Massachusetts Green High Performance Computing Center (MGHPCC). Each run took roughly 3-5 hours to complete.

\section{Results and discussion}
\label{sec:results}
Similar to the DisruptionBench paper \citep{lucas2024disruptionbench}, we use AUC and F2 score to assess the performance of each augmentation strategy on the four cases specified in Section \ref{sec:background}, setting C-MOD as the new machine (see Figure \ref{fig:machine-comparison} for an explanation of this choice). F2 score is used since false negatives are more costly in the case of disruption avoidance. The results are summarized in Table \ref{tab:results}.

\begin{table}
\caption{DisruptionBench results for LSTMFormer, FCN, and GPT-2 trained on using no augmentations, tsaug standard augmentations, and viewmaker augmentations during training. "New Machine" is C-MOD in each case. Refer to Section \ref{sec:background} for further descriptions of each case}
  \label{tab:results}
  \centering
\resizebox{\textwidth}{!}{
  \begin{tabular}{llcccccc}\toprule
      & & \multicolumn{2}{c}{LSTMFormer}& \multicolumn{2}{c}{FCN} & \multicolumn{2}{c}{GPT-2}\\
      \cmidrule(lr){3-4} \cmidrule(lr){5-6} \cmidrule(lr){7-8}
      & Aug Type & AUC & F2 & AUC & F2 & AUC & F2\\
      \midrule
      \multirow{3}{*}{Case 1: Zero-Shot} & None & 0.571& \textbf{0.499}& 0.518& \textbf{0.505}& 0.575& 0.525\\
      & TS & 0.571& 0.358& \textbf{0.561}& 0.166& \textbf{0.592}& \textbf{0.535}\\
      & Viewmaker & \textbf{0.609}& 0.342& 0.546& 0.503& 0.527& 0.404\\
      \midrule
      \multirow{3}{*}{Case 2: Few-Shot} & None & \textbf{0.592}& 0.389& 0.631& \textbf{0.513}& 0.695& 0.523\\
      & TS & 0.537& 0.189& 0.579& 0.413& 0.723& 0.616\\
      & Viewmaker & 0.568& \textbf{0.545}& \textbf{0.637}& 0.494& \textbf{0.731}& \textbf{0.628}\\
      \midrule
      \multirow{3}{*}{Case 3: All machines} & None & 0.808& 0.708& 0.847& 0.758& 0.711& 0.524\\
      & TS & 0.769& 0.641& 0.810& 0.72& 0.74& 0.58\\
      & Viewmaker & \textbf{0.823}& \textbf{0.735}& \textbf{0.853}& \textbf{0.777}& \textbf{0.754}& \textbf{0.611}\\
      \midrule
      \multirow{3}{*}{Case 4: Only CMOD} & None & \textbf{0.792}& \textbf{0.707}& 0.791& 0.7& 0.742& 0.617\\
      & TS & 0.79& 0.696& 0.781& 0.691& 0.675& 0.512\\
      & Viewmaker & 0.766& 0.669& \textbf{0.814}& \textbf{0.725}& \textbf{0.744}& \textbf{0.632}\\
      \midrule
      \multirow{3}{*}{Mean} & None & 0.691& \textbf{0.576}& 0.697& 0.619& 0.681& 0.547\\
      & TS & 0.667& 0.471& 0.683& 0.498& 0.683& 0.561\\
      & Viewmaker & \textbf{0.692}& 0.573& \textbf{0.713}& \textbf{0.625}& \textbf{0.689}& \textbf{0.569}\\
      \bottomrule
  \end{tabular}
}
\end{table}

\textbf{Better disruptive shot prediction}: As shown in Table \ref{tab:results}, the use of viewmaker views resulted in the highest average AUC scores across all post-hoc models with an improvement of 0.14\% for LSTMFormer, 2.3\% for FCN, and 1.17\% and for GPT-2. Views also contributed to the highest mean F2 score for FCN and GPT-2 with increases of 0.97\% and 4.02\% respectively. We also note that for case 3 (many shot case) we see a average improvement of 7.6\% in F2, and for case 2 (few shot case) we see a significant improvement of 18.1\%. On the other hand, tsaug seems to lag behind the viewmaker on average and actually drops mean F2 substantially for LSTMFormer (-18.23\%) and FCN (-19.55\%).  

\textbf{In-domain Augmentations}: The results on cases 2 and 3 in particular suggests that viewmakers help models learn underlying disruptive features across all machines. We hypothesize this is because viewmakers generate more physically-realistic augmentations of the original data. To explore this further, we compared the dynamic time warping (DTW) similarity of 250 randomly sampled disruptive discharges from each machine against their views and tsaugs (see Figure \ref{fig:dtw-comparison}). For this sample, the average DTW for views was 409.14 while for tsaugs it was 770.39, indicating that views are much more faithful to the original data. Additionally, we performed a Wilcoxon Signed Rank Test to get a $p<0.001$, showing this result is statistically significant.

\textbf{Faster disruption avoidance:} To understand how a model trained on views might behave in a production system, we plot a confusion matrix using one sampled shot from each category: TP, FP, TN, FN (Figure \ref{fig:disruptivity_plot}). As we can see, there are cases when the various augmentation strategies diverge. However, it seems that the views help models predict disruptions well in advance of the actual disruption, which raises the possibility of using these augmentations in training controllers for disruption avoidance.

\section{Limitations}
\label{sec:limitations}
Throughout the different post hoc models, training with viewmaker augmentations can cause loss in precision: -2.3\% in FCN, and -9.4\% in GPT-2; however, for LSTMFormer we see a +5.2\% improvement (see Table  \ref{tab:results-full}). The similarity of many disruptive and non-disruptive shots could be one reason for this tradeoff (see Figure \ref{fig:cmod-d-nd-comparison}) while class imbalance might be another. While ideally recall and precision would both increase, for disruption avoidance control the cost of false negatives significantly outweigh the costs of false positives. 
Finally, our access to compute was limited, so we did not have the ability to present runs on multiple seeds, which would strengthen the statistical significance of our results and allow us to test different combinations of data-splits and new machines.

\section{Conclusion and future work}

Avenues for future work can include modifying viewmaker features, integrating the trained encoder into a new classifier, and comparing viewmakers to other time series generators such as TimeGAN \citep{TimeGAN2019}. Additionally, more post-hoc models should be tested (see \citep{diiidRea, eastHu, asdexDis, zhuHybridDeeplearningArchitecture2020}), and new tokamaks should be added to DisruptionBench.

In this work, we propose a novel time series variant of viewmaker networks and investigate its utility in generating data augmentations or "views" of tokamak discharge time series to improve model robustness for nuclear fusion disruption prediction and disruption avoidance control. Our findings suggest that using viewmaker views improve model AUC and F2 across a diverse set prediction tasks compared to naive or no augmentations, paritcularly in few and many shot cases. By improving the robustness of these models, we move closer to making tokamaks a commercial and tangible realization of decades of work in the physical sciences.  

\pagebreak

\bibliographystyle{unsrtnat}
\bibliography{paper}

\begin{thebibliography}{40}
\providecommand{\natexlab}[1]{#1}
\providecommand{\url}[1]{\texttt{#1}}
\expandafter\ifx\csname urlstyle\endcsname\relax
  \providecommand{\doi}[1]{doi: #1}\else
  \providecommand{\doi}{doi: \begingroup \urlstyle{rm}\Url}\fi

\bibitem[Giuliani et~al.(2023)Giuliani, Grazian, Alotto, Agostini, Bustreo, and Zollino]{GIULIANI2023113554}
U.~Giuliani, S.~Grazian, P.~Alotto, M.~Agostini, C.~Bustreo, and G.~Zollino.
\newblock Nuclear {{Fusion}} impact on the requirements of power infrastructure assets in a decarbonized electricity system.
\newblock \emph{Fusion Engineering and Design}, 192:\penalty0 113554, 2023.
\newblock ISSN 0920-3796.
\newblock \doi{10.1016/j.fusengdes.2023.113554}.

\bibitem[Spangher et~al.(2019)Spangher, Vitter, and Umstattd]{spangher2019characterizing}
Lucas Spangher, J~Scott Vitter, and Ryan Umstattd.
\newblock Characterizing fusion market entry via an agent-based power plant fleet model.
\newblock \emph{Energy Strategy Reviews}, 26:\penalty0 100404, 2019.
\newblock \doi{10.1016/j.esr.2019.100404}.

\bibitem[Rea et~al.(2019)Rea, Montes, Erickson, Granetz, and Tinguely]{diiidRea}
Christina Rea, KJ~Montes, KG~Erickson, RS~Granetz, and RA~Tinguely.
\newblock A real-time machine learning-based disruption predictor in {DIII-D}.
\newblock \emph{Nuclear Fusion}, 59\penalty0 (9):\penalty0 096016, 2019.
\newblock \doi{10.1088/1741-4326/ab28bf/}.

\bibitem[Hu et~al.(2021)Hu, Rea, Yuan, Erickson, Chen, Shen, Huang, Xiao, Chen, Duan, et~al.]{eastHu}
WH~Hu, Cristina Rea, QP~Yuan, KG~Erickson, DL~Chen, Biao Shen, Yao Huang, JY~Xiao, JJ~Chen, YM~Duan, et~al.
\newblock Real-time prediction of high-density east disruptions using random forest.
\newblock \emph{Nuclear Fusion}, 61\penalty0 (6):\penalty0 066034, 2021.
\newblock \doi{10.1088/1741-4326/abf74d}.

\bibitem[Pautasso et~al.(2002)Pautasso, Tichmann, Egorov, Zehetbauer, Gruber, Maraschek, Mast, Mertens, Perchermeier, Raupp, et~al.]{asdexDis}
G~Pautasso, Ch~Tichmann, S~Egorov, T~Zehetbauer, O~Gruber, M~Maraschek, K-F Mast, V~Mertens, I~Perchermeier, G~Raupp, et~al.
\newblock On-line prediction and mitigation of disruptions in asdex upgrade.
\newblock \emph{Nuclear Fusion}, 42\penalty0 (1):\penalty0 100, 2002.
\newblock \doi{10.1088/0029-5515/42/1/314}.

\bibitem[Zhu et~al.(2020)Zhu, Rea, Montes, Granetz, Sweeney, and Tinguely]{zhuHybridDeeplearningArchitecture2020}
J.~X. Zhu, C.~Rea, K.~Montes, R.~S. Granetz, R.~Sweeney, and R.~A. Tinguely.
\newblock Hybrid deep-learning architecture for general disruption prediction across multiple tokamaks.
\newblock \emph{Nuclear Fusion}, 61\penalty0 (2):\penalty0 026007, December 2020.
\newblock ISSN 0029-5515.
\newblock \doi{10.1088/1741-4326/abc664}.

\bibitem[Arnold et~al.(2023)Arnold, Spangher, and Rea]{arnold2023continuous}
William~F Arnold, Lucas Spangher, and Christina Rea.
\newblock Continuous convolutional neural networks for disruption prediction in nuclear fusion plasmas, 2023.

\bibitem[Spangher et~al.(2023)Spangher, Arnold, Spangher, Maris, and Rea]{spangher2023autoregressive}
Lucas Spangher, William Arnold, Alexander Spangher, Andrew Maris, and Cristina Rea.
\newblock Autoregressive transformers for disruption prediction in nuclear fusion plasmas, 2023.

\bibitem[Granetz et~al.(2018)Granetz, Rea, Montes, Tinguely, Eidietis, Meneghini, Chen, Shen, Xiao, ERICKSON, et~al.]{granetz2018machine}
RS~Granetz, C~Rea, K~Montes, R~Tinguely, N~Eidietis, O~Meneghini, DL~Chen, B~Shen, BJ~Xiao, K~ERICKSON, et~al.
\newblock Machine learning for disruption warning on {Alcator C-Mod, DIII-D, and EAST} tokamaks.
\newblock In \emph{Proc. 27th IAEA Fusion Energy Conference, IAEA, Vienna}, 2018.
\newblock \doi{10.1088/1741-4326/ab1df4}.

\bibitem[Kallenbach et~al.(2017)Kallenbach, Team, Team, et~al.]{asdexUpgrade}
A~Kallenbach, ASDEX~Upgrade Team, EUROfusion~MST1 Team, et~al.
\newblock Overview of asdex upgrade results.
\newblock \emph{Nuclear Fusion}, 57\penalty0 (10):\penalty0 102015, 2017.
\newblock \doi{10.1088/1741-4326/aa64f6}.

\bibitem[Greenwald et~al.(2014)Greenwald, Bader, Baek, Bakhtiari, Barnard, Beck, Bergerson, Bespamyatnov, Bonoli, Brower, et~al.]{cMod}
Martin Greenwald, A~Bader, S~Baek, M~Bakhtiari, He~Barnard, W~Beck, W~Bergerson, I~Bespamyatnov, P~Bonoli, D~Brower, et~al.
\newblock 20 years of research on the alcator c-mod tokamak.
\newblock \emph{Physics of Plasmas}, 21\penalty0 (11), 2014.

\bibitem[Iwana and Uchida(2021)]{Iwana2021}
Brian~Kenji Iwana and Seiichi Uchida.
\newblock An empirical survey of data augmentation for time series classification with neural networks.
\newblock \emph{PLOS ONE}, 16\penalty0 (7):\penalty0 e0254841, July 2021.
\newblock ISSN 1932-6203.
\newblock \doi{10.1371/journal.pone.0254841}.
\newblock URL \url{http://dx.doi.org/10.1371/journal.pone.0254841}.

\bibitem[Hong and Travis(2022)]{YangRobust}
Yang Hong and Desell Travis.
\newblock Robust augmentation for multivariate time series classification.
\newblock 2022.
\newblock URL \url{https://doi.org/10.48550/arXiv.2201.11739}.

\bibitem[Um et~al.(2017)Um, Pfister, Pichler, Endo, Lang, Hirche, Fietzek, and Kulić]{Um_2017}
Terry~T. Um, Franz M.~J. Pfister, Daniel Pichler, Satoshi Endo, Muriel Lang, Sandra Hirche, Urban Fietzek, and Dana Kulić.
\newblock Data augmentation of wearable sensor data for parkinson’s disease monitoring using convolutional neural networks.
\newblock In \emph{Proceedings of the 19th ACM International Conference on Multimodal Interaction}, ICMI ’17. ACM, November 2017.
\newblock \doi{10.1145/3136755.3136817}.
\newblock URL \url{http://dx.doi.org/10.1145/3136755.3136817}.

\bibitem[Wen and Keyes(2019)]{wen2019time}
Tailai Wen and Roy Keyes.
\newblock Time series anomaly detection using convolutional neural networks and transfer learning, 2019.

\bibitem[Fu et~al.(2020)Fu, Kirchbuchner, and Kuijper]{fu2020data}
Biying Fu, Florian Kirchbuchner, and Arjan Kuijper.
\newblock Data augmentation for time series: traditional vs generative models on capacitive proximity time series.
\newblock In \emph{Proceedings of the 13th ACM international conference on pervasive technologies related to assistive environments}, pages 1--10, 2020.

\bibitem[Katharina et~al.(2022)Katharina, David, Bernd, Udo, Cristina, Andrew, Robert, and Christopher]{dpredaug22}
Rath Katharina, Rügamer David, Bischl Bernd, von~Toussaint Udo, Rea Cristina, Maris Andrew, Granetz Robert, and Albert Christopher.
\newblock Data augmentation for disruption prediction via robust surrogate models.
\newblock 2022.
\newblock URL \url{https://doi.org/10.1017/S0022377822000769}.

\bibitem[Tamkin et~al.(2021)Tamkin, Wu, and Goodman]{tamkin2021viewmaker}
Alex Tamkin, Mike Wu, and Noah Goodman.
\newblock Viewmaker networks: Learning views for unsupervised representation learning, 2021.

\bibitem[Lucas et~al.(2024)Lucas, Bonotto, Arnold, Chayapathy, Gallingani, Spangher, Cannarile, Bigoni, De~Marchi, and Rea]{lucas2024disruptionbench}
Spangher Lucas, Matteo Bonotto, William Arnold, Dhruva Chayapathy, Tommaso Gallingani, Alexander Spangher, Francesco Cannarile, Daniele Bigoni, Eliana De~Marchi, and Cristina Rea.
\newblock Disruptionbench: A robust benchmarking framework for machine learning-driven disruption prediction.
\newblock 2024.

\bibitem[Goodfellow et~al.(2014)Goodfellow, Pouget-Abadie, Mirza, Xu, Warde-Farley, Ozair, Courville, and Bengio]{goodfellow2014generative}
Ian~J. Goodfellow, Jean Pouget-Abadie, Mehdi Mirza, Bing Xu, David Warde-Farley, Sherjil Ozair, Aaron Courville, and Yoshua Bengio.
\newblock Generative adversarial networks, 2014.

\bibitem[Chen et~al.(2020{\natexlab{a}})Chen, Kornblith, Norouzi, and Hinton]{chen2020simple}
Ting Chen, Simon Kornblith, Mohammad Norouzi, and Geoffrey Hinton.
\newblock A simple framework for contrastive learning of visual representations, 2020{\natexlab{a}}.

\bibitem[Wu et~al.(2022)Wu, Xu, Wang, and Long]{wu2022autoformer}
Haixu Wu, Jiehui Xu, Jianmin Wang, and Mingsheng Long.
\newblock Autoformer: Decomposition transformers with auto-correlation for long-term series forecasting, 2022.

\bibitem[Zeyer et~al.(2019)Zeyer, Bahar, Irie, Schlüter, and Ney]{zeyer2019compare}
Albert Zeyer, Parnia Bahar, Kazuki Irie, Ralf Schlüter, and Hermann Ney.
\newblock A comparison of transformer and lstm encoder decoder models for asr.
\newblock In \emph{2019 IEEE Automatic Speech Recognition and Understanding Workshop (ASRU)}, pages 8--15, 2019.
\newblock \doi{10.1109/ASRU46091.2019.9004025}.

\bibitem[Lim et~al.(2020)Lim, Arik, Loeff, and Pfister]{lim2020temporal}
Bryan Lim, Sercan~O. Arik, Nicolas Loeff, and Tomas Pfister.
\newblock Temporal fusion transformers for interpretable multi-horizon time series forecasting, 2020.

\bibitem[Ba et~al.(2016)Ba, Kiros, and Hinton]{ba2016layer}
Jimmy~Lei Ba, Jamie~Ryan Kiros, and Geoffrey~E. Hinton.
\newblock Layer normalization, 2016.

\bibitem[Vaswani et~al.(2017)Vaswani, Shazeer, Parmar, Uszkoreit, Jones, Gomez, Kaiser, and Polosukhin]{vaswani2017attention}
Ashish Vaswani, Noam Shazeer, Niki Parmar, Jakob Uszkoreit, Llion Jones, Aidan~N. Gomez, Lukasz Kaiser, and Illia Polosukhin.
\newblock Attention is all you need, 2017.

\bibitem[Wang et~al.(2016)Wang, Yan, and Oates]{wang2016timeseriesclassificationscratch}
Zhiguang Wang, Weizhong Yan, and Tim Oates.
\newblock Time series classification from scratch with deep neural networks: A strong baseline, 2016.
\newblock URL \url{https://arxiv.org/abs/1611.06455}.

\bibitem[Lucas et~al.(2023)Lucas, William, Alexande, Andrew, and Cristina]{lucasgpt-2}
Spangher Lucas, Arnold William, Spangher Alexande, Maris Andrew, and Rea Cristina.
\newblock Autoregressive transformers for disruption prediction in nuclear fusion plasmas.
\newblock 2023.
\newblock URL \url{https://doi.org/10.48550/arXiv.2401.00051}.

\bibitem[Yoon et~al.(2019)Yoon, Jarrett, and van~der Schaar]{TimeGAN2019}
Jinsung Yoon, Daniel Jarrett, and Mihaela van~der Schaar.
\newblock Time-series generative adversarial networks.
\newblock In H.~Wallach, H.~Larochelle, A.~Beygelzimer, F.~d\textquotesingle Alch\'{e}-Buc, E.~Fox, and R.~Garnett, editors, \emph{Advances in Neural Information Processing Systems}, volume~32. Curran Associates, Inc., 2019.
\newblock URL \url{https://proceedings.neurips.cc/paper_files/paper/2019/file/c9efe5f26cd17ba6216bbe2a7d26d490-Paper.pdf}.

\bibitem[Bachman et~al.(2019)Bachman, Hjelm, and Buchwalter]{bachman2019learning}
Philip Bachman, R~Devon Hjelm, and William Buchwalter.
\newblock Learning representations by maximizing mutual information across views, 2019.

\bibitem[He et~al.(2020)He, Fan, Wu, Xie, and Girshick]{he2020momentum}
Kaiming He, Haoqi Fan, Yuxin Wu, Saining Xie, and Ross Girshick.
\newblock Momentum contrast for unsupervised visual representation learning, 2020.

\bibitem[Chen et~al.(2020{\natexlab{b}})Chen, Fan, Girshick, and He]{chen2020improved}
Xinlei Chen, Haoqi Fan, Ross Girshick, and Kaiming He.
\newblock Improved baselines with momentum contrastive learning, 2020{\natexlab{b}}.

\bibitem[Gao et~al.(2022)Gao, Yao, and Chen]{gao2022simcse}
Tianyu Gao, Xingcheng Yao, and Danqi Chen.
\newblock Simcse: Simple contrastive learning of sentence embeddings, 2022.

\bibitem[Yue et~al.(2022)Yue, Wang, Duan, Yang, Huang, Tong, and Xu]{yue2022ts2vec}
Zhihan Yue, Yujing Wang, Juanyong Duan, Tianmeng Yang, Congrui Huang, Yunhai Tong, and Bixiong Xu.
\newblock Ts2vec: Towards universal representation of time series, 2022.

\bibitem[Goodfellow et~al.(2015)Goodfellow, Shlens, and Szegedy]{goodfellow2015explaining}
Ian~J. Goodfellow, Jonathon Shlens, and Christian Szegedy.
\newblock Explaining and harnessing adversarial examples, 2015.

\bibitem[Minderer et~al.(2020)Minderer, Bachem, Houlsby, and Tschannen]{minderer2020automatic}
Matthias Minderer, Olivier Bachem, Neil Houlsby, and Michael Tschannen.
\newblock Automatic shortcut removal for self-supervised representation learning, 2020.

\bibitem[Jernigan et~al.(2005)Jernigan, Baylor, Combs, Humphreys, Parks, and Wesley]{massiveGasInjection}
T.~C. Jernigan, L.~A. Baylor, S.~K. Combs, D.~A. Humphreys, P.~B. Parks, and J.~C. Wesley.
\newblock Massive gas injection systems for disruption mitigation on the diii-d tokamak.
\newblock In \emph{21st IEEE/NPS Symposium on Fusion Engineering SOFE 05}, pages 1--3, 2005.
\newblock \doi{10.1109/FUSION.2005.252977}.

\bibitem[Jachmich et~al.(2021)Jachmich, Kruezi, Lehnen, Baruzzo, Baylor, Carnevale, Craven, Eidietis, Ficker, Gebhart, et~al.]{jachmich2021shattered}
S~Jachmich, Uron Kruezi, Michael Lehnen, M~Baruzzo, Larry~R Baylor, D~Carnevale, Dylan Craven, NW~Eidietis, O~Ficker, TE~Gebhart, et~al.
\newblock Shattered pellet injection experiments at jet in support of the iter disruption mitigation system design.
\newblock \emph{Nuclear Fusion}, 62\penalty0 (2):\penalty0 026012, 2021.
\newblock \doi{10.1088/1741-4326/ac3c86}.

\bibitem[Keith et~al.(2024)Keith, Nagpal, Rea, and Tinguely]{keith2024risk}
Zander Keith, Chirag Nagpal, Cristina Rea, and R.A. Tinguely.
\newblock Risk-{{Aware Framework Development}} for {{Disruption Prediction}}: {{Alcator C-Mod}} and {{DIII-D Survival Analysis}}.
\newblock 2024.

\bibitem[Greenwald(2002)]{greenwald2002density}
Martin Greenwald.
\newblock Density limits in toroidal plasmas.
\newblock \emph{Plasma Physics and Controlled Fusion}, 44\penalty0 (8):\penalty0 R27, 2002.

\end{thebibliography}
\medskip

%%%%%%%%%%%%%%%%%%%%%%%%%%%%%% Appendix %%%%%%%%%%%%%%%%%%%%%%%%%%%%%
\appendix
\section{Viewmaker networks}
\label{app:viewmaker}
\subsection{Contrastive learning}
Contrastive learning is a technique which aims to help models learn an embedding space in which views derived from the same input are grouped together while dissimilar views are pushed apart. The success of contrastive learning approaches has been extensively researched, particularly on computer vision benchmarks \citep{bachman2019learning, he2020momentum, chen2020simple, chen2020improved}, but also in other domains \citep{gao2022simcse, yue2022ts2vec}. While generating handcrafted views for tasks such as image classification have been refined, the process remains a handicap for domains such as time series.

SimCLR loss is one approach to contrastive learning proposed in \citep{chen2020simple} for visual data. If given \(N\) pairs of views generated from the same input \(X\), the SimCLR loss is defined to be
\begin{equation}\label{equ:SimCLR}
    \begin{split}
    \mathcal{L} & = \frac{1}{2N} \sum_{k=1}^{N} [\ell(2k-1, 2k) + \ell(2k, 2k-1)], \\
    \ell(i, j) & = -\log\frac{\exp(s(i,j))}{\sum_{k=1}^{2N} \mathbbm{1}_{k \neq i} \exp(s(i, k))} \\
    \end{split}
\end{equation}
Here, \(s(i,j)\) is the cosine similarity of embeddings of views i and j.

\subsection{Viewmaker components}
\begin{figure}
    \centering
    \includegraphics[width=0.8\linewidth]{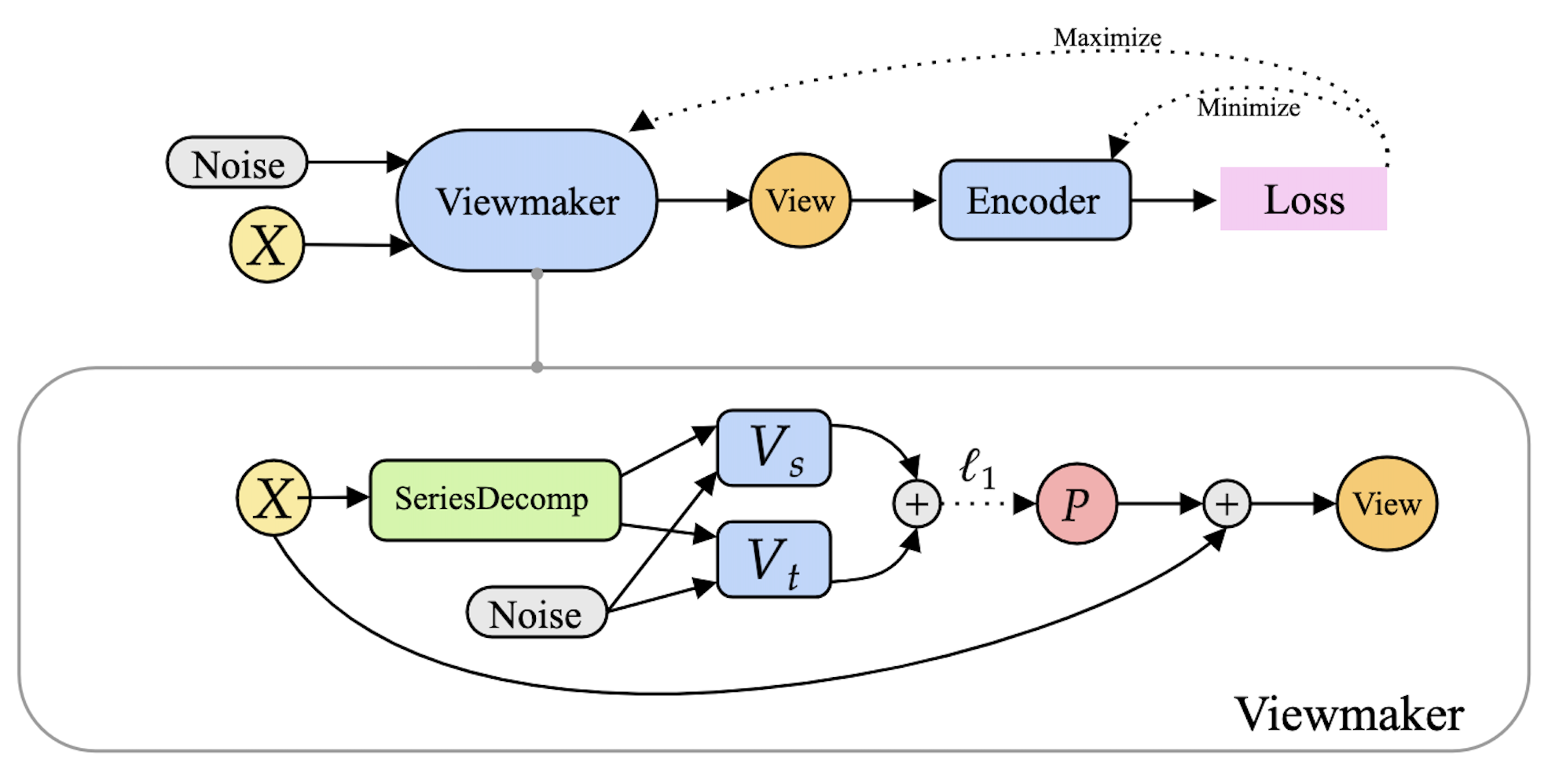}
    \caption{Overview of the time series viewmaker. \(V_t\) and \(V_s\) are generator networks which generate perturbed time series from \(X_t\) and \(X_s\) respectively.}
    \label{fig:TSViewmaker}
\end{figure}

To visually reinforce the brief description of viewmakers in Section \ref{sec:background}, we encourage readers to refer to the top illustration in Figure \ref{fig:TSViewmaker}, which gives a high-level overview of the components of a viewmaker network as described in \citep{tamkin2021viewmaker}. The bottom illustration highlights the specific changes to the viewmaker created for this work which is discussed in Section \ref{sec:models}.

In order to force the encoder to learn useful latent representations of the original data, the viewmaker network presents the encoder with stochastic perturbations of the original data projected onto an \(\ell_p\) ball. Such projections have been studied in the context of adversarial robustness (e.g. in \citep{goodfellow2015explaining}).
The strength of the perturbations is further controlled by a distortion-budget: a hyperparameter which limits the radius of the \(\ell_p\) ball to ensure that views don’t become too difficult to learn useful representations from.

The advantage to using viewmakers over handcrafted augmentations is their ability to learn augmentations from the data directly. In the context of disruption prediction, viewmakers could help models pick up hidden trends and physical characteristics of disruptive discharges \citep{minderer2020automatic} that are machine-agnostic: a critical step in the pursuit of more generalized models. 

\section{Background (continued)}
\subsection{DisruptionBench}
\label{app:disruptionbench}
\begin{figure}
    \centering
    \includegraphics[width=\linewidth]{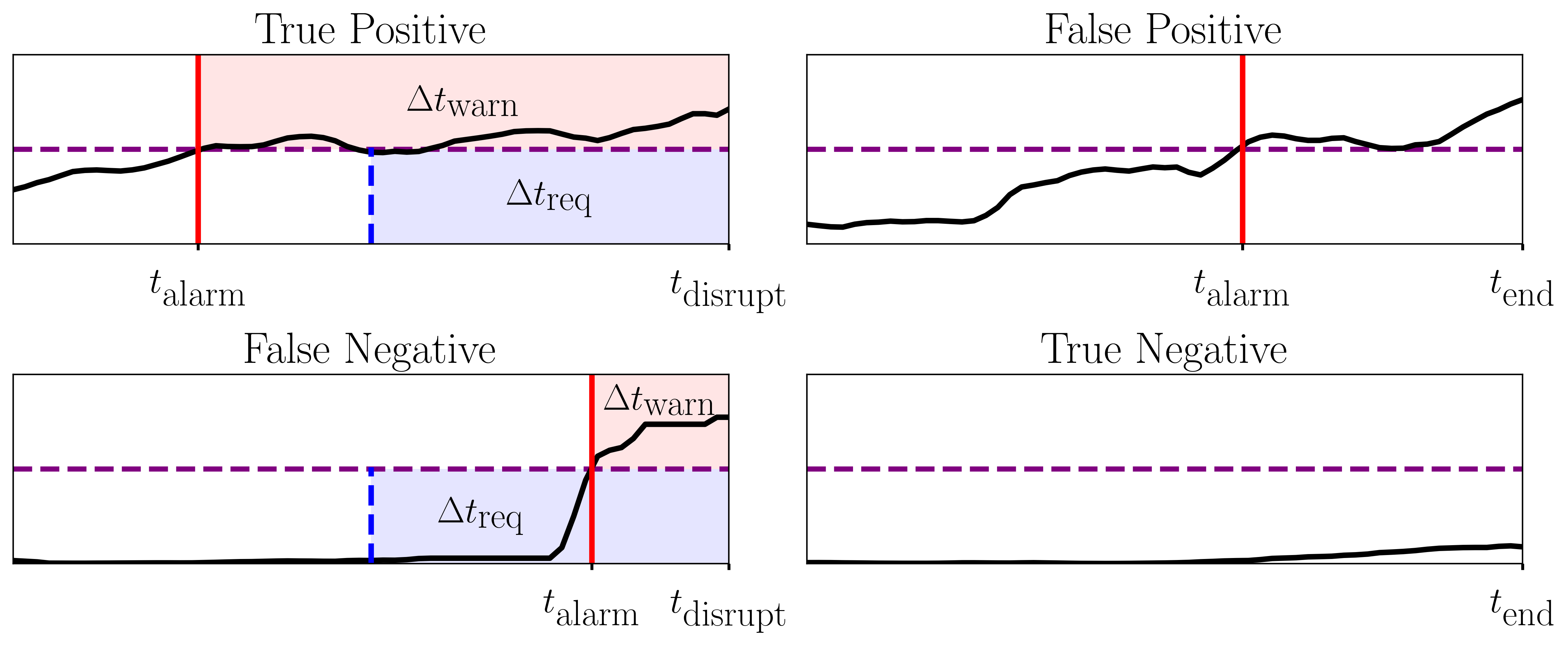} 
    \caption{Illustration of the categorization of true and false positives, and true and false negatives. Figure adapted from ref. \citep{keith2024risk}.}
    \label{fig:categorization_confusion_matrix}
\end{figure}

Designation of true or false positives in DisruptionBench is calculated similar to how a production-scale Disruption Mitigation System (DMS) would behave. The DMS needs at least some time to engage\footnote{Two common methods of mitigating the harm from a disruption include: (1) puffing a heavy inert gas like Neon or Argon around the chamber's walls \citep{massiveGasInjection} or (2) shooting a frozen hydrogen pellet into the middle of the plasma reaction \citep{jachmich2021shattered}. Both methods introduce particles that absorb the energy of the plasma.}, $\Delta t_{\text{req}}$. Two thresholds $T_{low}$, $T_{high} \in (0, 1)$ and a hysteresis value $h$ is set, and a prediction model outputs \textit{disruptivity} scores, i.e. the probability that a disruption will happen in the future. If the disruptivity is above $T_{high}$ for $h$ time steps before $\Delta t_{\text{req}}$, the prediction is categorized as positive at time $\Delta t_{\text{warn}}$. Else, the discharge reaches its planned end and the model predicts a negative. If the disruptivity dips below $T_{low}$, the hysteresis counter is reset. For an illustration of how this translates into true and false positives and negatives, please see Figure \ref{fig:categorization_confusion_matrix}. 

\begin{table}[!htb]
    \caption{The input features of the model, their definitions, and a categorization of the type of instability the signal indicates. The features listed are the same across all machines. While this table is adapted from \citep{zhuHybridDeeplearningArchitecture2020}, some of our symbology, features, and feature descriptions are unique.}
    \label{tab:features}
    \centering
    \renewcommand{\arraystretch}{1.45}
    \begin{tabular}{p{0.2\linewidth}p{.4\linewidth}p{0.15\linewidth}p{0.15\linewidth}}
        \toprule
        Feature & Definition & Relevant Instability & Units \\
        \midrule
        % Locked mode indicator & Locked mode magnetic field normalized to toroidal field & MHD & Unitless\\
        % Rotating mode indicator & Standard deviation of Mirnov array normalized by toroidal field & MHD & Unitless\\
        $\beta_p$ & Plasma pressure normalized by poloidal magnetic pressure & MHD & Unitless\\
        $\ell_i$ & Normalized plasma internal inductance & MHD & Unitless \\
        $q_{95}$ & Safety factor at 95\% of normalized flux surfaces & MHD & Unitless \\
        $n=1$ mode & $n=1$ component of the perturbed magnetic field & MHD & $T$ \\
        $n/n_G$ & Electron density normalized by Greenwald density limit \citep{greenwald2002density} & Dens. limit & Unitless \\
        % $\Delta z_{\textrm{center}}$ & Vertical position error of plasma current centroid & Vert. Stab. & $m$ \\
        Lower Gap & Gap between plasma and lower divertor & Shaping & $m$\\
        $\kappa$ & Plasma elongation & Shaping & Unitless \\
        % $P_{\textrm{rad}}/P_{\textrm{input}}$ & Radiated power normalized by input power & Impurities & $W/W$ \\
        $I_{\textrm{p,error}}/I_{\textrm{p,prog}}$ & Plasma current deviation from the programmed plasma trace (error), normalized by the programmed plasma trace (prog)  & Impurities, hardware, runaway electrons  & Unitless \\
        $V_{\text{loop}}$ & Toroidal ``loop'' voltage & Impurities & $V$ \\ 
        Tokamak Indicators & Three additional features with binary encodings of whether the data came from Alcator C-Mod, DIII-D, or EAST & n.a. & Unitless \\
        \bottomrule
    \end{tabular}
\end{table}

% put into appendix. 
\begin{table}[!ht]
    \caption{Metrics on a dataset composed of multiple tokamaks.}
    \label{tab:datasets}
    \centering
    \begin{tabular}{cccccc}
        \toprule
        Tokamak & $\tau$ & Number of Discharges & Average Discharge Length & Initial Sampling Rate \\
        \midrule
        C-Mod &  50 ms & 4000 & 0.52 s & 0.005 ms \\
        DIII-D & 150 ms & 8000 & 3.7 s & 0.01 ms \\
        EAST & 400 ms & 11000 & 5.3 s & 0.025 ms \\
        \bottomrule
    \end{tabular}
\end{table}

\section{Results (continued)}

%% Full table %%
\begin{table}
\caption{Full DisruptionBench results (including Recall and Precision) for LSTMFormer, FCN, and GPT-2 trained on using no augmentations, tsaug standard augmentations, and viewmaker augmentations during training. "New Machine" is C-MOD in each case.}
\label{tab:results-full}
\centering
\resizebox{\textwidth}{!}{%
\begin{tabular}{@{}llcccccccccccc@{}}
\toprule
& & \multicolumn{4}{c}{LSTMFormer}& \multicolumn{4}{c}{FCN}& \multicolumn{4}{c}{GPT-2}\\
\cmidrule(lr){3-6} \cmidrule(lr){7-10} \cmidrule(lr){11-14}
& Aug Type & AUC & Recall & Precision & F2 & AUC & Recall & Precision & F2 & AUC & Recall & Precision & F2\\
\midrule
\multirow{3}{*}[-1ex]{\makecell[l]{Case 1:\\ Zero-Shot}} 
& None & 0.571 & \textbf{0.772} & 0.207 & \textbf{0.499} & 0.518 & \textbf{0.912} & 0.181 & \textbf{0.505} & 0.575 & \textbf{0.860} & 0.205 & 0.525 \\
& TS & 0.571 & 0.404 & 0.247 & 0.358 & \textbf{0.561} & 0.140 & \textbf{0.615} & 0.166 & \textbf{0.592} & 0.860 & \textbf{0.213} & \textbf{0.535} \\
& Viewmaker & \textbf{0.609} & 0.333 & \textbf{0.380} & 0.342 & 0.546 & 0.842 & 0.193 & 0.503 & 0.527 & 0.561 & 0.190 & 0.404 \\
\midrule
\multirow{3}{*}[-1ex]{\makecell[l]{Case 2:\\ Few-Shot}} 
& None & \textbf{0.592} & 0.439 & 0.269 & 0.389 & 0.631 & \textbf{0.684} & 0.257 & \textbf{0.513} & 0.695 & 0.561 & \textbf{0.410} & 0.523 \\
& TS & 0.537 & 0.175 & \textbf{0.270} & 0.189 & 0.579 & 0.509 & 0.236 & 0.413 & 0.723 & 0.789 & 0.328 & 0.616 \\
& Viewmaker & 0.568 & \textbf{0.965} & 0.199 & \textbf{0.545} & \textbf{0.637} & 0.614 & \textbf{0.278} & 0.494 & \textbf{0.731} & \textbf{0.807} & 0.333 & \textbf{0.628} \\
\midrule
\multirow{3}{*}[-1ex]{\makecell[l]{Case 3:\\ All machines}} 
& None & 0.808 & 0.807 & 0.474 & 0.708 & 0.847 & 0.824 & \textbf{0.573} & 0.758 & 0.711 & 0.526 & 0.517 & 0.524 \\
& TS & 0.769 & 0.702 & \textbf{0.476} & 0.641 & 0.810 & 0.877 & 0.420 & 0.720 & 0.740 & 0.596 & \textbf{0.523} & 0.580 \\
& Viewmaker & \textbf{0.823} & \textbf{0.877} & 0.446 & \textbf{0.735} & \textbf{0.853} & \textbf{0.930} & 0.469 & \textbf{0.777} & \textbf{0.754} & \textbf{0.649} & 0.493 & \textbf{0.611} \\
\midrule
\multirow{3}{*}[-1ex]{\makecell[l]{Case 4:\\ Only CMOD}} 
& None & \textbf{0.792} & \textbf{0.912} & 0.371 & \textbf{0.707} & 0.791 & 0.877 & 0.388 & 0.700 & 0.742 & 0.719 & \textbf{0.394} & 0.617 \\
& TS & 0.790 & 0.860 & \textbf{0.395} & 0.696 & 0.781 & 0.877 & 0.373 & 0.691 & 0.675 & 0.579 & 0.351 & 0.512 \\
& Viewmaker & 0.766 & 0.842 & 0.366 & 0.669 & \textbf{0.814} & \textbf{0.877} & \textbf{0.427} & \textbf{0.725} & \textbf{0.744} & \textbf{0.772} & 0.367 & \textbf{0.632} \\
\midrule
\multirow{3}{*}[-1ex]{Mean} 
& None & 0.691 & 0.733 & 0.330 & \textbf{0.576} & 0.697 & \textbf{0.824} & 0.350 & 0.619 & 0.681 & 0.667 & \textbf{0.382} & 0.547 \\
& TS & 0.667 & 0.535 & 0.347 & 0.471 & 0.683 & 0.601 & \textbf{0.411} & 0.498 & 0.683 & \textbf{0.706} & 0.354 & 0.561 \\
& Viewmaker & \textbf{0.692} & \textbf{0.754} & \textbf{0.347} & 0.573 & \textbf{0.713} & 0.816 & 0.342 & \textbf{0.625} & \textbf{0.689} & 0.697 & 0.346 & \textbf{0.569} \\
\bottomrule
\end{tabular}%
}
\end{table}

%% Machine UMAP %%
\begin{figure}
    \centering
    \includegraphics[width=0.75\linewidth]{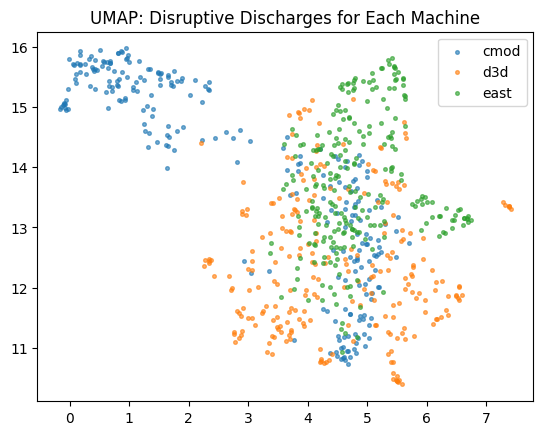}
    \caption{A UMAP clustering of disruptive discharges from each machine. Many discharges from C-MOD exhibit distinct behavior from those from DIII-D and EAST. Thus, to challenge models in their ability to learn general disruptive features, we designate C-MOD as the "new machine" parameter in our experiments.}
    \label{fig:machine-comparison}
\end{figure}

%% DTW comparison %%
\begin{figure}
    \centering
    \includegraphics[width=0.85\linewidth]{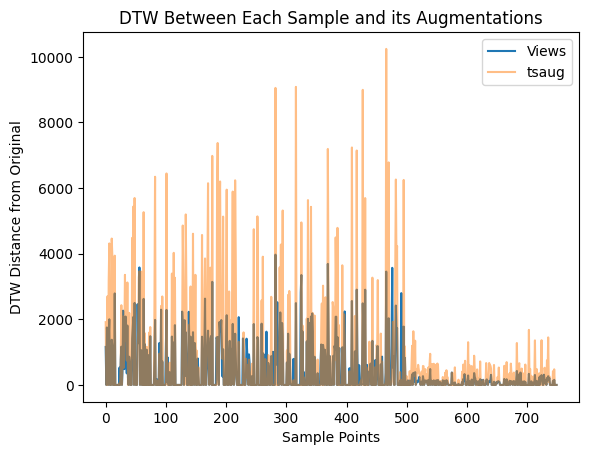}
    \caption{Comparison of dynamic time warping (DTW) similarities between disruptive discharges and their views (in cyan) and their tsaugs (in orange). Discharges were sampled evenly at random across each machine.}
    \label{fig:dtw-comparison}
\end{figure}

\begin{figure}[!htb]
    \centering
    \includegraphics[width=\textwidth]{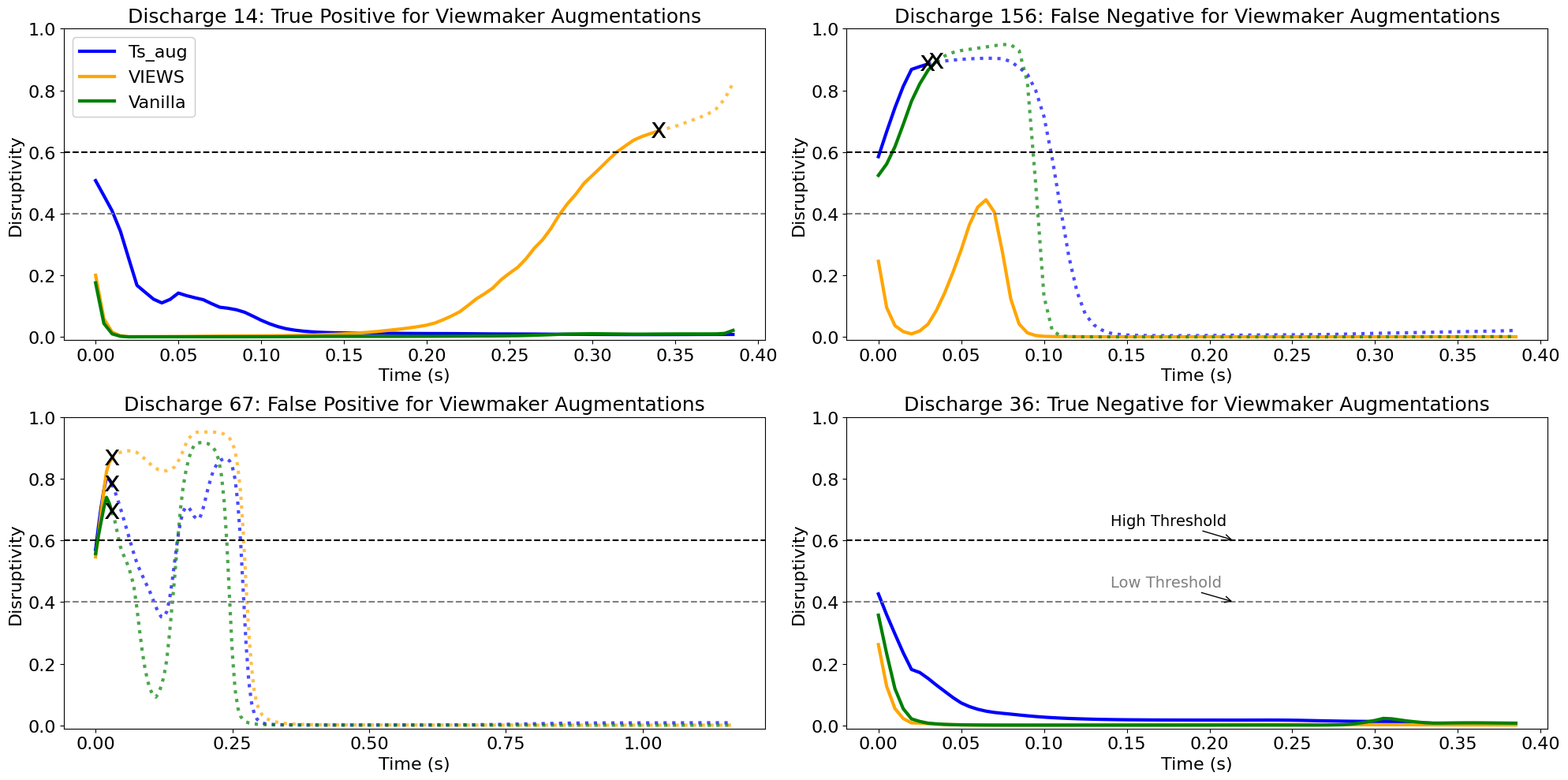}
    \caption{A comparison of four discharges from Alcator C-Mod for unrolled disruptivity on the test set, chosen based on outcomes of an LSTMFormer trained of views. We annotate the chart with ``X'''s to show the points at which the disruption models would have triggered the DMS. The disruptivity metric, on the y-axis, is unitless. Finally, note that the timebases are not true, but relative to the start of the recorded data.}
    \label{fig:disruptivity_plot}
\end{figure}

% D vs. ND for CMOD
\begin{figure}
    \centering
    \includegraphics[width=0.85\linewidth]{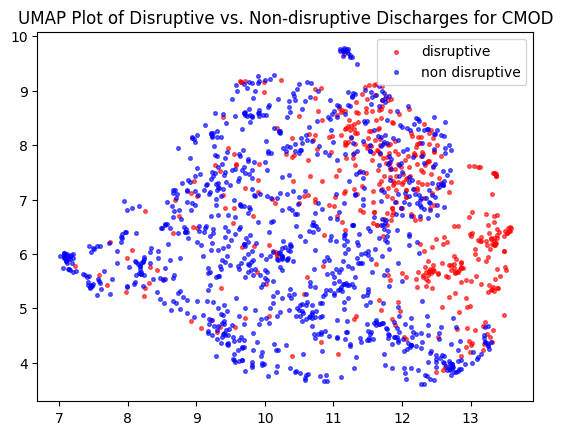}
    \caption{UMAP comparison of disruptive and non-disruptive discharges randomly sampled from Alcator C-MOD.}
    \label{fig:cmod-d-nd-comparison}
\end{figure}

\end{document}